\documentclass[conference,onecolumn]{IEEEtran}
\IEEEoverridecommandlockouts
\usepackage{cite}
\usepackage{amsmath,amssymb,amsfonts}
\usepackage{algorithmic}
\usepackage{graphicx}
\usepackage{textcomp}
\usepackage{xcolor}
\usepackage{float}
\usepackage{booktabs}
\usepackage{makecell}

\begin{document}

\title{ Deep Compression of Neural Networks for Fault Detection on Tennessee Eastman Chemical Processes}

\author{Mingxuan Li \\
\IEEEauthorblockA{\textit{Department of Computer Science} \\
\textit{University of North Carolina at Chapel Hill}\\
Chapel Hill, NC, USA \\
mingxuan\_li@unc.edu}

\and
\IEEEauthorblockN{* Yuanxun Shao}
\IEEEauthorblockA{
Nuro Inc.\\
Mountain View, CA, USA \\
yuanxun@gatech.edu}
}

\maketitle
\thispagestyle{plain}
\pagestyle{plain}

\begin{abstract}
Artificial neural network has achieved the state-of-art performance in fault detection on the Tennessee Eastman process, but it often requires enormous memory to fund its massive parameters. In order to implement online real-time fault detection, three deep compression techniques (pruning, clustering, and quantization) are applied to reduce the computational burden. We have extensively studied 7 different combinations of compression techniques, all methods achieve high model compression rates over 64\% while maintain high fault detection accuracy. The best result is applying all three techniques, which reduces the model sizes by 91.5\% and remains a high accuracy over 94\%.  This result leads to a smaller storage requirement in production environments, and makes the deployment smoother in real world.
\end{abstract}


\section{Introduction}
Artificial Neural Network (ANN) is a powerful technique that can classify input data into predefined categories. ANN has been successfully deployed in an enormous amount of real-life applications with superior results, including computer vision \cite{Ciresan:2012}, natural language processing \cite{Socher:2013}, autonomous driving \cite{Chen:2015}. For a classification problem, input data is processed through a series of weight matrices, bias vectors, and nonlinear activation functions. Following by a classifier, ANN eventually calculates a likelihood score for each category. During training, ANN maximizes the likelihood of the true category (or minimize the negative of this likelihood) for taring data sets via variants of stochastic gradient decent methods. In terms of the inference, we run the forward path and classify testing data to the category with the highest probability score. Even though ANN has shown great success in many fields, sometimes the enormous model size caused by a large number of weights can result in overwhelming computational burden (e.g. taking huge memory and being very power demanding). In certain application scenarios that have memory or energy consumption limitations, such as mobile platform or browser-based systems, a smaller size ANN with high accuracy is not only necessary but also hold in high demand.

Deep compression is a powerful technique for reducing the size of an ANN while remaining high accuracy. There are three techniques can be used to achieve this goal: pruning, clustering, and quantization. Pruning is the process of removing relatively insignificant weights from weight matrices. Since the matrices can be stored in the sparse matrix data type, the removal of any item in one matrix directly reduces its size. Clustering, also known as weight sharing, is the process of using the same value for multiple weights. Since one reference can be used for several weights using this technique, it can reduce the storage space for storing weights. At last, quantization is the technique of storing weights in data types with lower precision in exchange for smaller space in storage. By combing those three techniques, we are able to reduce the size of ANN while maintaining a similar accuracy.

Fault detection, as an essential step of industrial production to ensure efficiency and safety, has been an active field of research for past years. As the complexity of production increases dramatically due to more advanced technologies, deep learning has become an increasingly popular candidate for fault detection applications due to its potential in handling complex systems, including a large amount of measurements and many failure types for processes. In this paper, we focus on the Tennessee Eastman process, a representative case for fault detection and diagnostic. The Tennessee Eastman process is a typical industrial process that consists of five main process units. It has 52 measurements and 21 fault types as defined by Downs and Vogel in \cite{Downs:1993}. Moreover, as a widely used case for fault detection, the usage of ANN to diagnose fault has been proved to have satisfying results. To achieve the best accuracy, instead of using a single ANN to classify the fault type, the result in \cite{Heo:2018} uses different ANNs for different fault types and achieves 97.73\% average accuracy. However, this method requires one deep learning model for each fault type, which requires large storage space and computational efforts in total. Tennessee Eastman process requires the online detection ability with low latency. Thus, the ANN pruning is an ideal candidate for reducing the computational burden, speeding up the online inference time, as well as maintaining the classification accuracy.

In this paper, deep compression of artificial neural networks (ANN) is used for the fault detection on Tennessee Eastman chemical process to enable faster online inferences with high accuracy. Sections \ref{sec: DL} and \ref{sec: DC} introduce the basics of deep learning and deep compression. Section \ref{sec: FD} discusses the details of deep learning architecture and deep compression results for fault detection of TN Eastman chemical processes. Finally, Section \ref{sec: conclusion} summaries this paper.


\section{deep learning}
\label{sec: DL}
An ANN usually has 4 different types of layer: input layer, hidden layer, softmax layer, and output layer. Input layer is a weight matrix designed to pass the input data into the ANN through a linear mapping. In Hidden layers, the input information is transformed through a matrix multiplication and an activation function: $h_1 = \sigma(W_1x+b_1)$, $h_i = \sigma(W_ih_{i-1}+b_i), i = \{2, 3, 4, ..., k\}$, where $x\in\mathbb{R}^{n_x}$ is the vector of input data, $h_i\in\mathbb{R}^{n_{h_i}}$ is the $i_{th}$ hidden layer representation, $W_i\in \mathbb{R}^{n_{h_i}\times{n_{h_{i-1}}}}$ and $b_i\in \mathbb R^{n_{h_i}}$ are weights and bias for connecting the  i-th and  (i-1)-th hidden representations, $k$ is the number of hidden layers, and $\sigma$ is a nonlinear activation function that introduces nonlinearity into the neural network. Specifically, ReLU activation function, $\sigma(x) = max(0,x)$, is used in this paper. The last fully-connected hidden layer does not impose the activation function. Then, a softmax layer calculates the score/probability of each class via $\frac{e^{f_{y_i}}}{\sum_je^{f_j}}$, where $f_{y_i}$ is the score of the correct category, and $f_j$ is the score of the $j_{th}$ catagory.  Finally, the loss is calculated based on the true label and regularization. The objective of the network is to minimize the loss and maximum the accuracy of both training and testing datasets.

\begin{figure}[H]

\centering 
\includegraphics[width=0.7\textwidth]{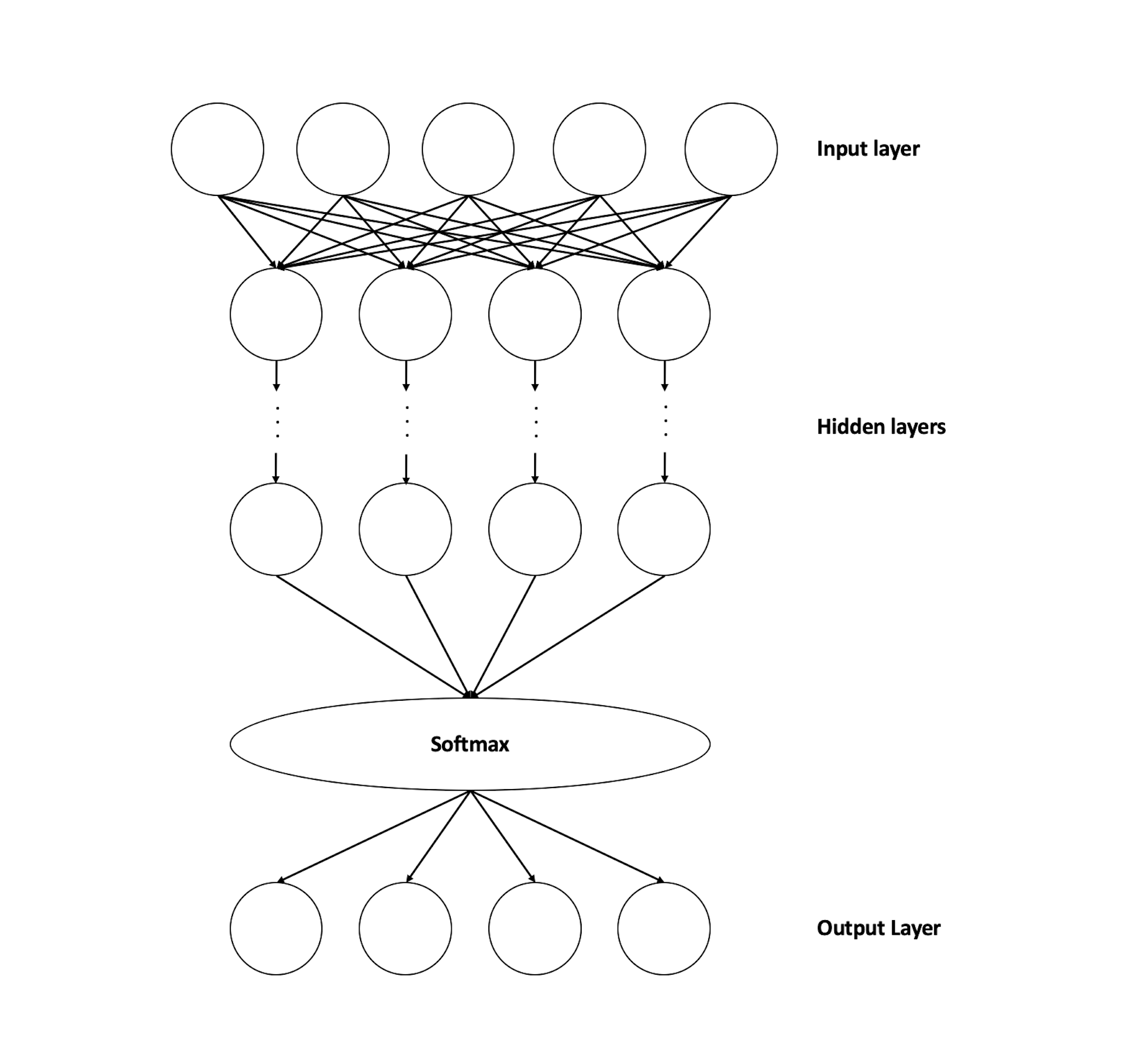} 
\caption{ANN Structure} 
\label{Figure 1} 
\end{figure}

\section{Deep Compression}
\label{sec: DC}
Neural networks usually require enormous memory to fund their massive parameters. In this section, three deep compression techniques are discussed to reduce the model size and keep high accuracy. 

\subsection{Pruning}

\begin{figure}[H]
\centering 
\includegraphics[width=0.45\textwidth]{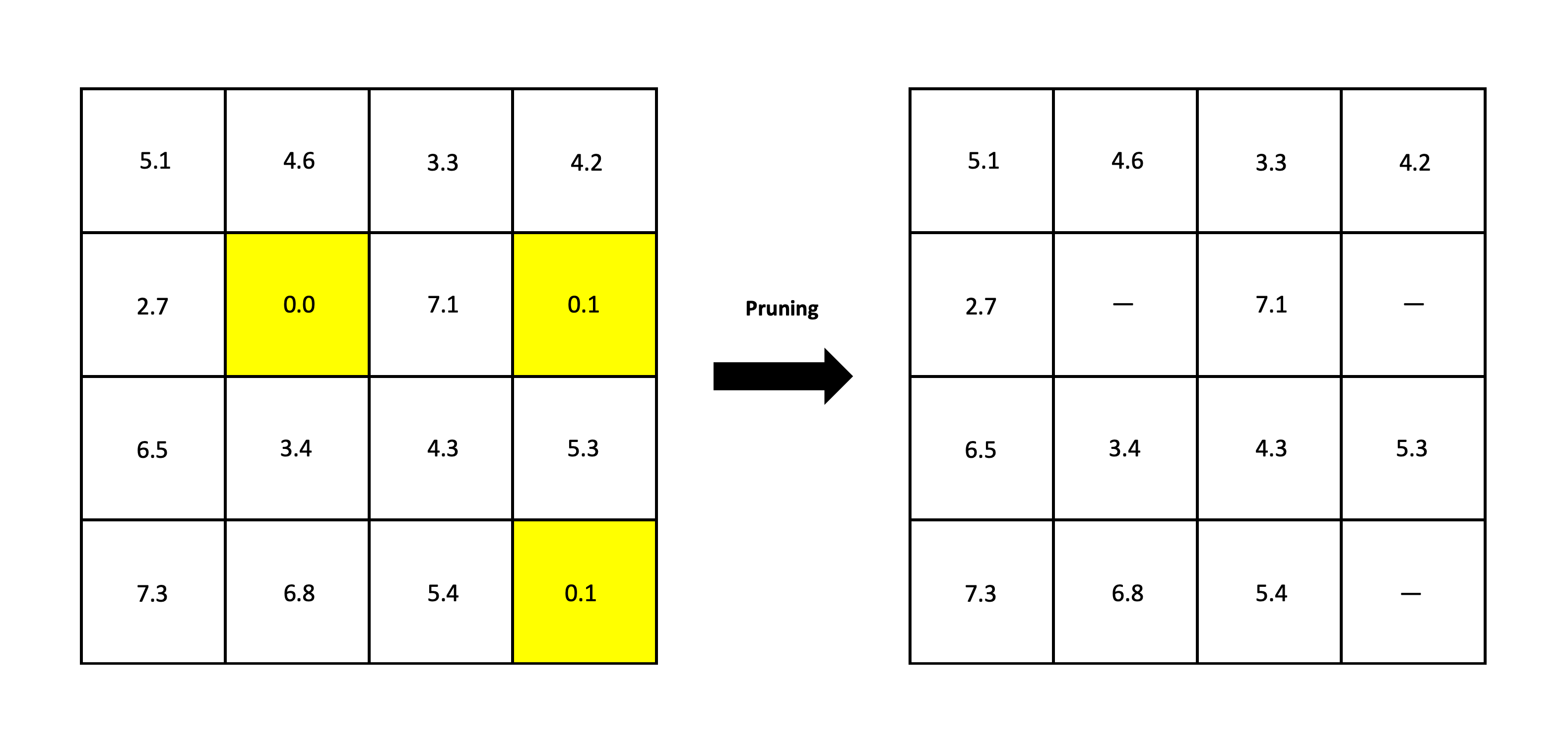} 
\caption{The left figure is the original weight matrix, and the right figure is the matrix after pruning with a threshold 0.1. Highlighted weights are removed.} 
\label{Figure 2} 
\end{figure}

Network pruning utilizes the connectivity between neuron s. Small weights under a predefined threshold are removed to reduce the model size. As shown in Figure \ref{Figure 2}, the left figure is the original weights matrix. After pruning with a threshold 0.1, the pruned weights are illustrated in the right figure. The sparsity of the matrix reduces model size, and can speeds up the calculation with hardware optimizations. 

\subsection{Clustering}
\begin{figure}[H]
\centering 
\includegraphics[width=0.65\textwidth]{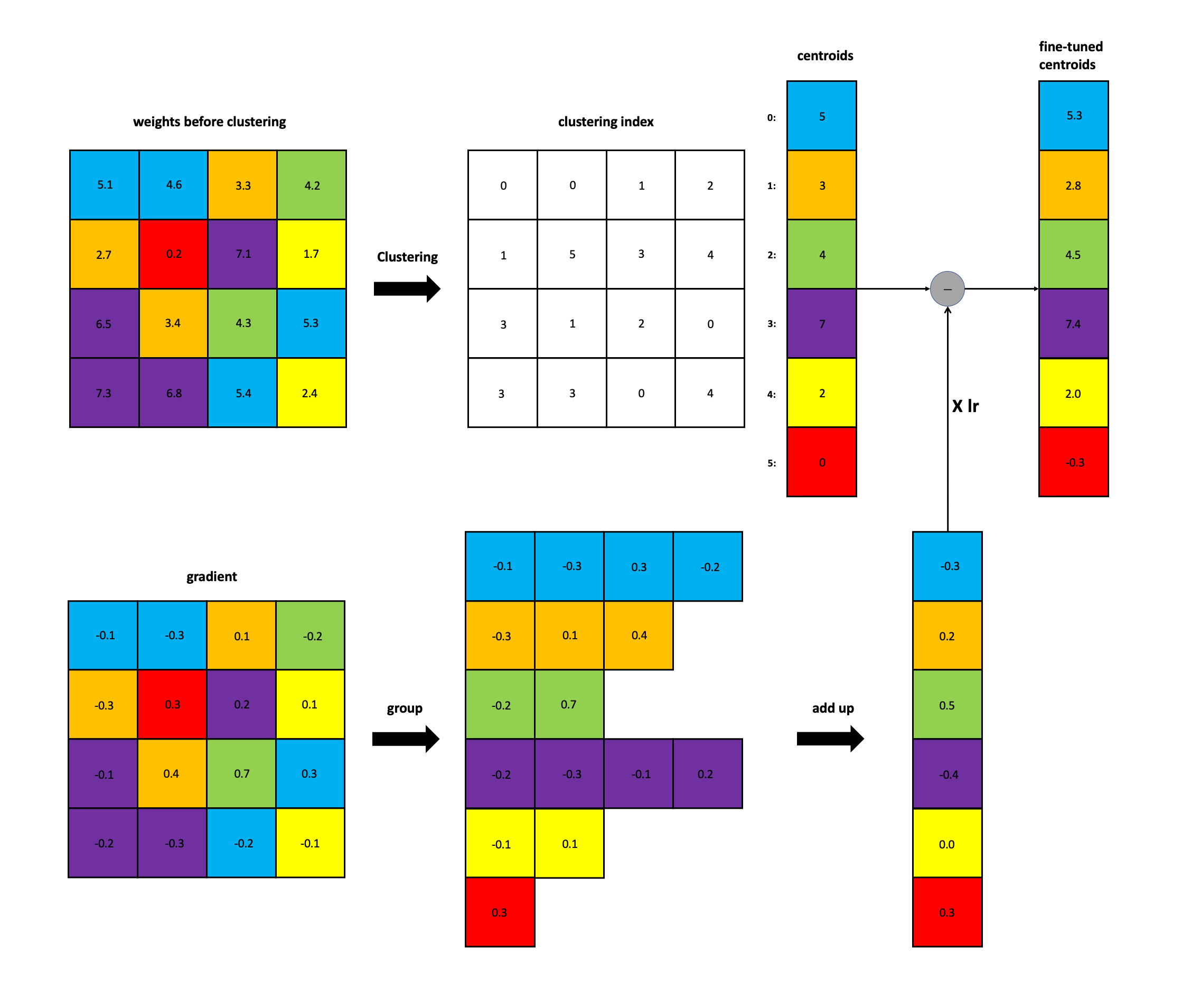} 
\caption{An illustration of the gradient update for clustered weights: the top left figure is the original weight matrix colored according to clustered groups; the top middle figure is the corresponding group indices; the bottom left figure is the gradient of the weight matrix; the bottom middle figure is the grouped gradients according to the centroids; the bottom right figure shows the sum of gradients of each group; and the top right figure represent the final gradient update of clusters.} 
\label{Figure 3} 
\end{figure}

Clustering is a technique that groups weights with close values. First, weights are divided into a fixed number of groups. Second, values in the same group will be reset as the same value and thus all weights in the same group can be referenced to the same address to reduce model size. In Figure \ref{Figure 3}, the left top figure is a weight matrix before clustering. Weights in this original matrix are grouped into 6 groups, and all weights in the same group are each assigned with a same value. The gradient of each weight is calculated through back-propagation and the gradient of each group is the sum over all elements. Finally, weights are updated through the cluster groups. Clustering essentially minimizes the amount of references needed for a model, and thus reduces the storage requirement of this model.

\subsection{Quantization}
\begin{figure}[H]
\centering 
\includegraphics[width=0.7\textwidth]{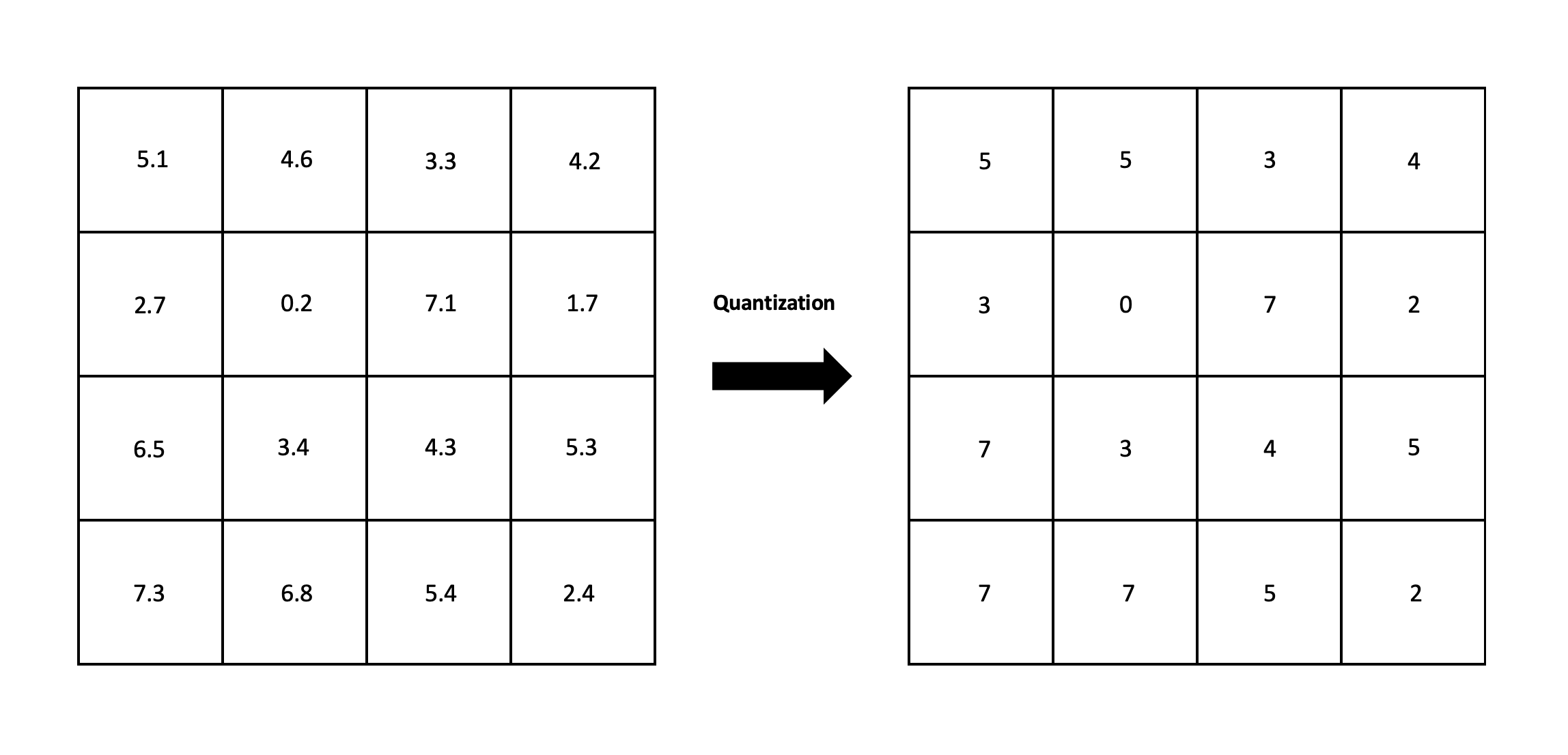} 
\caption{The left figure is the weight matrix before quantization; the right figure is the weight matrix after quantization.} 
\label{Figure 4} 
\end{figure}

In ANN, weights are usually stored a in high precision format such as float64 or float32, which consumes a relatively large space. In the quantization process, the precision used to store weights is degraded in the exchange for a lower space consumption. In Figure \ref{Figure 4}, weights in the left figure are stored in one decimal, while the right figure reduces the storage requirement via rounding off the decimal value.

\section{Fault Detection for the Tennessee Eastman Chemical Process}
\label{sec: FD}
The Tennessee Eastman process is a benchmark for fault detection \cite{Heo:2018}. Based on a baseline ANN structure, various combinations of deep compression techniques are applied to reduce the model size and achieve a high accuracy.

\subsection{Tennessee Eastman Process Description}
As summarized in Table \ref{Table I}, the Tennessee Eastman process provides 52 measurements and 21 different fault types (including "no fault" as faultNumber 0) in its dataset. Moreover, Chiang \cite{Chiang:2001} and Zhang \cite{Zhang:2009} pointed out that due to the absence of observable change in the mean, variance and the higher order variances in the data, we do not have enough information to detect Fault 3, 9, and 15. Therefore, this paper follows the common rule to exclude the three faults above for the consideration, resulting in 18 classification types (including "no failure" as faultNumber 0).

\begin{table}[H]   
\centering
\begin{tabular}{lcc}    
\toprule Fault Number&Process Variable & Type \\    
\midrule   
faultNumber 1 & A/C feed ratio, B composition con- stant (stream 4)  & Step \\
faultNumber 2 & B composition, A/C ratio constant (stream 4) & Step \\
faultNumber 3 & D feed temperature (stream 2) & Step \\
faultNumber 4 & Reactor cooling water inlet tem- perature&Step \\
faultNumber 5 & Condenser cooling water inlet tem- perature&Step\\
faultNumber 6 & A feed loss (stream 1)&Step\\ faultNumber 7 & C header pressure loss - reduced availability (stream 4)&Step\\
faultNumber 8 & A, B, C feed composition (stream 4)&Random variation\\
faultNumber 9 &  D feed temperature (stream 2)&Random variation\\
faultNumber 10 & C feed temperature (stream 4)
&Random variation\\
faultNumber 11 & Reactor cooling water inlet tem- perature&Random variation\\
faultNumber 12 & Condenser cooling water inlet tem- perature&Random variation\\
faultNumber 13 & Reaction kinetics&Slow drift\\
faultNumber 14 & Reactor cooling water valve&Sticking\\
faultNumber 15 & Condenser cooling water valve&Sticking\\
faultNumber 16 & Unknown& Unkown\\
faultNumber 17 & Unknown& Unkown\\
faultNumber 18 & Unknown& Unkown\\
faultNumber 19 & Unknown& Unkown\\
faultNumber 20 & Unknown& Unkown\\
\bottomrule   
& &\\
\end{tabular}  
\caption{Fault Types in the TE Process}  
\label{Table I}
\end{table}

\subsection{Baseline Artificial Neural Network Models}
In Tennessee Eastman process measurements, one entry in the dataset might correspond to multiple fault types. Following the state-of-art method in \cite{Heo:2018}, we construct and train 18 different ANNs to detect each of the 18 fault types. In baseline, each ANN has 6 layers in the structure of $52-64-256-128-256-128-64-2$. The input layer dimension is 52, which corresponds to the 52 measurement. The output layer has a dimension of 2, which corresponds to the 2 possible outcomes (fault or no fault). The resulting 18 models are served as the baseline for comparison with compressed models.

\subsection{Deep Compression Results}
In this section, all 7 different combinations of pruning, clustering, and quantization are applied to the 18 ANNs. First, each of the three compression techniques is applied individually to exam the performance. Then, each two of the compression techniques applied to the original 18 ANNs. At Last, all three techniques are applied to the original ANNs to compare the compression rate and accuracy change.

The compression results of ANNs obtained by applying each of the three techniques individually are shown in Table \ref{Table II}, \ref{Table III}, and \ref{Table IV}. As shown in Table \ref{Table II}, when only pruning is applied, the average compressed rate is 64.0\% , and the average accuracy change is -0.5\% for all 18 different types of fault labels. This result is promising considering the average accuracy only changed slightly, while the average model size is significantly reduced by 64.0\%. Table \ref{Table III} shows the compression result after clustering only, which achieves 76.0\% average compressed rate and only -1.2\% average accuracy change. This method is the best among combinations that only apply a single compression technique. Table \ref{Table IV} presents the result after only quantization, which achieves an average compression rate of 72.0\%, and an average accuracy change of -7.8\%.

\begin{table}[H]   
\centering
\scalebox{0.85}{
\begin{tabular}{lcccccc}    
\toprule Fault & \makecell[c]{ANN \\ Size (bytes)} & ANN Acc (\%) & \makecell[c]{Compressed \\ Size (bytes)} & \makecell[c]{Compressed \\Acc (\%)}  & \makecell[c]{Compressed\\ Rate (\%)} & Acc Change (\%)\\     
\midrule   
0 & 486924  & 94.4 & 175207 & 94.4 & 64.0 & 0.0\\

1 & 487120  & 99.4 & 175235 & 95.3 & 64.0 & -4.1 \\

2 & 486853  & 99.6 & 174988 & 99.6 & 64.1 & 0.0 \\

3 & 486931  & 96.1 & 175159 & 94.4 & 64.0 & -1.7\\

4 & 486901  & 95.2 & 175354  & 95.3 & 64.0 & 0.1  \\

5 & 486753  & 95.6 & 174983 & 99.8 & 64.1 & 4.2\\

6 & 486914  & 99.8 & 174765  & 99.8 & 64.1 & 0.0\\

7 & 486992  & 97.9 & 175556  & 98.5 & 64.0 & 6.0\\

8 & 486789  & 95.3 & 175412  & 95.3 & 64.0 & 0.0 \\

9 & 486782  & 94.4 & 175086  & 94.4 & 64.0 & 0.0 \\

10 & 486737  & 95.3 & 175564  & 95.3 & 64.0 & 0.0\\

11 & 487124  & 97.6 & 175382  & 97.3 & 64.0 & -0.3\\

12 & 486838  & 96.4 & 175283  & 99.1 & 64.0 & 2.7 \\

13 & 486819  & 97.1 & 175111  & 95.3 & 64.0 & -1.8\\

14 & 486844  & 96.8 & 175266  & 96.8 & 64.0 & 0.0\\

15 & 486899  & 96.4 & 175400  & 92.0 & 64.0 & -4.4\\

16 & 486816  & 95.3 & 175605  & 94.4 & 64.0 & -0.9\\

17 & 486844  & 97.6 & 175296  & 94.4 & 64.0 & -3.2\\

\bottomrule   
Average &  &  &  &  & 64.0 & -0.5\\
\end{tabular}  
}
\caption{Compression results with pruning} 
\label{Table II}
\end{table}

\begin{table}[H]   
\centering
\scalebox{0.85}{
\begin{tabular}{lcccccc}    
\toprule Fault & \makecell[c]{ANN \\ Size (bytes)} & ANN Acc (\%) & \makecell[c]{Compressed \\ Size (bytes)} & \makecell[c]{Compressed \\Acc (\%)}  & \makecell[c]{Compressed\\ Rate (\%)} & Acc Change (\%)\\        
\midrule   

0 & 486924  & 94.4 & 117313  & 94.4 & 75.9 & 0.0 \\

1 & 487120  & 99.4 & 117725  & 99.1 & 75.8 & -0.3\\

2 & 486853  & 99.6 & 116627  & 99.4 & 76.0 & -0.2\\

3 & 486931  & 96.1 & 118057  & 94.4 & 75.8 & -1.7\\

4 & 486901  & 95.2 & 116922  & 94.4 & 76.0 & -0.8 \\

5 & 486753  & 95.6 & 115861  & 99.6 & 76.2 & 4.0\\

6 & 486914  & 99.8 & 114344  & 94.4 & 76.5 & -5.4\\

7 & 486992  & 97.9 & 115685  & 94.9  & 76.2 & -3.0\\

8 & 486789  & 95.3 & 116442  & 94.4 & 76.1 & -0.9\\

9 & 486782  & 94.4 & 117599  & 94.4 & 75.8 & 0.0\\

10 & 486737  & 95.3 & 117798  & 94.4 & 75.8 & -0.9\\

11 & 487124  & 97.6 & 115934  & 94.7 & 76.2 & -2.9\\

12 & 486838  & 96.4 & 117672  & 94.4  & 75.8 & -2.0\\

13 & 486819  & 97.1 & 116671  & 94.4 & 76.0 & -2.7\\

14 & 486844  & 96.8 & 118284  & 96.3 & 75.7 & -0.5\\

15 & 486899  & 96.4 & 116166  & 94.7  & 76.1 & -1.7\\

16 & 486816  & 95.3 & 117600  & 94.4 & 75.8 & -0.9\\

17 & 486844  & 97.6 & 117014  & 94.5 & 76.0 & -3.1\\
\bottomrule   
Average &  &  &  &  & 76.0 & -1.2\\
\end{tabular}  
}
\caption{Compression results with clustering}  
\label{Table III}
\end{table}

\begin{table}[H]   
\centering
\scalebox{0.85}{
\begin{tabular}{lcccccc}    
\toprule Fault & \makecell[c]{ANN \\ Size (bytes)} & ANN Acc (\%) & \makecell[c]{Compressed \\ Size (bytes)} & \makecell[c]{Compressed \\Acc (\%)}  & \makecell[c]{Compressed\\ Rate (\%)} & Acc Change (\%)\\        
\midrule   

0 & 486924  & 94.4 & 137620  & 94.4 & 71.7 & 0.0 \\

1 & 487120  & 99.4 & 137030  & 95.9 & 71.9 & -3.5\\

2 & 486853  & 99.6 & 137482  & 98.8 & 71.8 & -0.8\\

3 & 486931  & 96.1 & 136678  & 94.4 & 71.9 & -1.7\\

4 & 486901  & 95.2 & 137766  & 94.4 & 71.7 & -0.8 \\

5 & 486753  & 95.6 & 137140  & 99.5 & 71.8 & 3.9\\

6 & 486914  & 99.8 & 136784  & 13.7 & 71.9 & -86.1\\

7 & 486992  & 97.9 & 137008  & 94.8  & 71.9 & -3.1\\

8 & 486789  & 95.3 & 137479  & 94.4 & 71.8 & -0.9\\

9 & 486782  & 94.4 & 137911  & 94.4 & 71.7 & 0.0\\

10 & 486737  & 95.3 & 137582  & 94.4 & 71.7 & -0.9\\

11 & 487124  & 97.6 & 137075  & 95.5 & 71.9 & -2.1\\

12 & 486838  & 96.4 & 137428  & 71.8  & 75.8 & -25.6\\

13 & 486819  & 97.1 & 137064  & 82.9 & 71.8 & -14.2\\

14 & 486844  & 96.8 & 137421  & 95.8 & 71.8 & -1.0\\

15 & 486899  & 96.4 & 137109  & 96.1  & 71.8 & -0.3\\

16 & 486816  & 95.3 & 137560  & 94.4 & 71.7 & -0.9\\

17 & 486844  & 97.6 & 137238  & 95.0 & 71.8 & -2.6\\
\bottomrule   
Average &  &  &  &  & 72.0 & -7.8\\
\end{tabular}  
}
\caption{Compression results with Quantization}  
\label{Table IV}
\end{table}

Next, a comprehensive study of applying two compression techniques are conducted. Table \ref{Table V} shows that when utilizing both clustering and pruning, the average compressed rate increases to 87.3\%, and the average accuracy change is only -1.9\%. Moreover, the compression result is consistent among all fault types. The variance of compressed rate and accuracy change are respectively 0.16 and 1.72, which shows that there are few fluctuations in compression rate and accuracy change, thus the average compression rate and average accuracy change are reliably representative. Table \ref{Table VI} shows the compression result with both pruning and quantization. The average compression rate is 88.1\%, which is about the same as the result in Table \ref{Table V}. But the average accuracy change drops to -5.3\%, which is the largest drop among methods that apply two compression techniques. Table \ref{Table VII} presents the compression result with both clustering and pruning. The average accuracy change is -1.8\%, which is close to the average accuracy change in Table \ref{Table V}. The compression rate is 82.2\%, which is 5.1\% smaller than the result in Table \ref{Table V}. In general, applying two compression techniques achieves better results than only applying one single technique.

\begin{table}[H]   
\centering
\scalebox{0.9}{
\begin{tabular}{lcccccc}    
\toprule  Fault & \makecell[c]{ANN \\ Size (bytes)} & ANN Acc (\%) & \makecell[c]{Compressed \\ Size (bytes)} & \makecell[c]{Compressed \\Acc (\%)}  & \makecell[c]{Compressed\\ Rate (\%)} & Acc Change (\%)\\      
\midrule   
0 & 486924  & 94.4 & 62014  & 94.4 & 87.3 & 0.0\\

1 & 487120  & 99.4 & 61788  & 95.3 & 87.3 & -4.1 \\

2 & 486853  & 99.6 & 63234  & 95.3 & 87.0 & -4.3\\

3 & 486931  & 96.1 & 61694  & 94.4 & 87.3 & -1.7\\

4 & 486901  & 95.2 & 62078  & 95.3 & 87.3 & 0.1\\

5 & 486753  & 95.6 & 62995  & 95.3 & 87.0 & -0.3 \\

6 & 486914  & 99.8 & 62065  & 95.3 & 87.3 & -4.5\\

7 & 486992  & 97.9 & 60908  & 95.3 & 87.5 & -2.6 \\

8 & 486789  & 95.3 & 60772  & 95.3 & 87.5 & 0.0 \\

9 & 486782  & 94.4 & 61664  & 94.4 & 87.3 & 0.0\\

10 & 486737  & 95.3 & 62178  & 95.3 & 87.2 & 0.0\\

11 & 487124  & 97.6 & 61671  & 95.3 & 87.3 & -2.3\\

12 & 486838  & 96.4 & 62292  & 95.3 & 87.2 & 0.0\\

13 & 486819  & 97.1 & 61002  & 95.3 & 87.5 & -1.8\\

14 & 486844  & 96.8 & 62443  & 92.1 & 87.2 & -4.7\\

15 & 486899  & 96.4 & 60988  & 94.4 & 87.5 & -2.0\\

16 & 486816  & 95.3 & 60709  & 94.4 & 87.5 & -0.9\\

17 & 486844  & 97.6 & 61086  & 94.4 & 87.5 & -3.2\\
\bottomrule  
Average &  &  &  &  & 87.3 & -1.9
\end{tabular}  
}
\caption{Compression results with Pruning and Clustering}  
\label{Table V}
\end{table}

\begin{table}[H]   
\centering
\scalebox{0.9}{
\begin{tabular}{lcccccc}    
\toprule  Fault & \makecell[c]{ANN \\ Size (bytes)} & ANN Acc (\%) & \makecell[c]{Compressed \\ Size (bytes)} & \makecell[c]{Compressed \\Acc (\%)}  & \makecell[c]{Compressed\\ Rate (\%)} & Acc Change (\%)\\      
\midrule   
0 & 486924  & 94.4 & 57823  & 94.4 & 88.1 & 0.0\\

1 & 487120  & 99.4 & 58337  & 94.4 & 88.0 & -5.0 \\

2 & 486853  & 99.6 & 57901  & 94.2 & 88.1 & -5.4\\

3 & 486931  & 96.1 & 58162  & 94.4 & 88.1 & -1.7\\

4 & 486901  & 95.2 & 58014  & 94.4 & 88.1 & -0.8\\

5 & 486753  & 95.6 & 57901  & 94.4 & 88.1 & -1.2 \\

6 & 486914  & 99.8 & 58132  & 94.4 & 88.1 & -5.4\\

7 & 486992  & 97.9 & 57978  & 94.4 & 88.1 & -3.5 \\

8 & 486789  & 95.3 & 57330  & 94.4 & 88.2 & -0.9 \\

9 & 486782  & 94.4 & 58013  & 94.4 & 88.1 & 0.0\\

10 & 486737  & 95.3 & 58098  & 94.4 & 88.1 & -0.9\\

11 & 487124  & 97.6 & 57775  & 94.4 & 88.1 & -3.2\\

12 & 486838  & 96.4 & 58371  & 94.4 & 88.0 & -2.0\\

13 & 486819  & 97.1 & 56579  & 94.4 & 88.4 & -2.7\\

14 & 486844  & 96.8 & 58048  & 46.9 & 88.1 & -49.9\\

15 & 486899  & 96.4 & 57806  & 87.1 & 88.1 & -9.3\\

16 & 486816  & 95.3 & 57919  & 94.4 & 88.1 & -0.9\\

17 & 486844  & 97.6 & 57648  & 94.4 & 88.2 & -3.2\\
\bottomrule  
Average &  &  &  &  & 88.1 & -5.3
\end{tabular}  
}
\caption{Compression results with Pruning and Quantization}  
\label{Table VI}
\end{table} 

\begin{table}[H]   
\centering
\scalebox{0.9}{
\begin{tabular}{lcccccc}    
\toprule  Fault & \makecell[c]{ANN \\ Size (bytes)} & ANN Acc (\%) & \makecell[c]{Compressed \\ Size (bytes)} & \makecell[c]{Compressed \\Acc (\%)}  & \makecell[c]{Compressed\\ Rate (\%)} & Acc Change (\%)\\      
\midrule   
0 & 486924  & 94.4 & 86997  & 94.4 & 82.1 & 0.0\\

1 & 487120  & 99.4 & 86233  & 96.6 & 82.3 & -2.8 \\

2 & 486853  & 99.6 & 86674  & 95.7 & 82.3 & -3.9\\

3 & 486931  & 96.1 & 85853  & 94.4 & 82.4 & -1.7\\

4 & 486901  & 95.2 & 86820  & 94.4 & 82.2 & -0.8\\

5 & 486753  & 95.6 & 86477  & 99.2 & 82.2 & 3.6 \\

6 & 486914  & 99.8 & 85627  & 94.4 & 82.4 & -5.4\\

7 & 486992  & 97.9 & 86582  & 94.5 & 82.2 & -3.4 \\

8 & 486789  & 95.3 & 86966  & 94.4 & 82.1 & -0.9 \\

9 & 486782  & 94.4 & 86958  & 94.4 & 82.1 & 0.0\\

10 & 486737  & 95.3 & 86962  & 94.3 & 82.1 & -1.0\\

11 & 487124  & 97.6 & 86509  & 93.7 & 82.2 & -3.9\\

12 & 486838  & 96.4 & 86261  & 94.4 & 82.3 & -2.0\\

13 & 486819  & 97.1 & 86494  & 94.4 & 82.2 & -2.7\\

14 & 486844  & 96.8 & 86644  & 94.7 & 82.2 & -2.1\\

15 & 486899  & 96.4 & 86561  & 94.3 & 82.2 & -2.1\\

16 & 486816  & 95.3 & 86990  & 94.4 & 82.1 & -0.9\\

17 & 486844  & 97.6 & 86580  & 94.4 & 82.2 & -3.2\\
\bottomrule  
Average &  &  &  &  & 82.2 & -1.8
\end{tabular}  
}
\caption{Compression results with Clustering and Quantization}  
\label{Table VII}
\end{table} 

Finally, we exam the performance of applying all three compression techniques. Table \ref{Table VIII} shows that the average compressed rate increases to 91.5\% while the average accuracy change remains relatively small at -1.8\%. The variances of the compressed rates and accuracy changes are only 0.14 and 1.83, correspondingly, which shows the consistency of the performance among all fault types. Compare to results of applying two techniques in Table \ref{Table V}, Table \ref{Table VI}, and Table \ref{Table VII}, this method achieves the best compressed rate with no extra loss in accuracy, which indicates applying all three techniques is better than only applying two techniques for fault diagnosis in the TE process. Compared to the results of applying only one compression technique in Table \ref{Table II}, \ref{Table III}, and \ref{Table IV}, although average accuracy change decreases slightly, applying all three techniques still achieves accuracies higher than 94\% for all fault types. At the same time, the compressed rate 91.5\% is significantly higher than 76\%, the best result of applying only one technique. 

\begin{figure}[H]
\centering 
\includegraphics[width=1.0\textwidth]{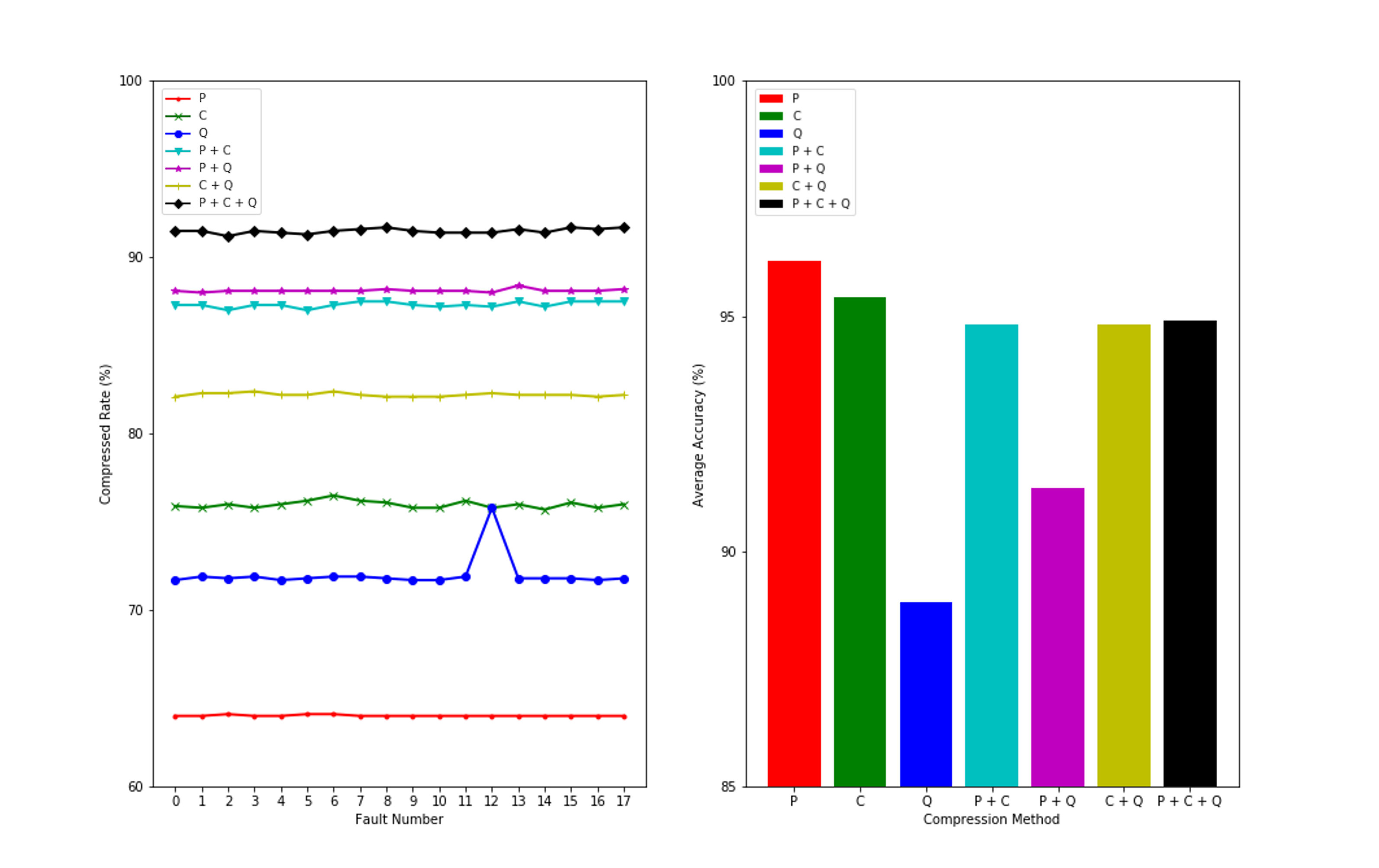} 
\caption{The left plot shows compressed rates of all fault types over 7 different combinations of compression techniques, where P = pruning, C = clustering, and Q = quantization. The right plot shows the average accuracy over all fault types with 7 different combinations of compression techniques. Results of different compression methods are colored consistently in two plots.} 
\label{Figure 5} 
\end{figure}

The results of all 7 different combination of compression techniques are summarized in Figure \ref{Figure 5}. It is clear that applying all three techniques achieves the highest compressed rate, as well as maintains a high average accuracy above 94\%. Thus, it is ideal to utilize all three compression techniques for the fault diagnosis in the TE process.

\begin{table}[H]   
\centering
\scalebox{0.9}{
\begin{tabular}{lcccccc}    
\toprule Fault & \makecell[c]{ANN \\ Size (bytes)} & ANN Acc (\%) & \makecell[c]{Compressed \\ Size (bytes)} & \makecell[c]{Compressed \\Acc (\%)}  & \makecell[c]{Compressed\\ Rate (\%)} & Acc Change (\%)\\        
\midrule   

0 & 486924  & 94.4 & 41391  & 94.4 & 91.5 & 0.0 \\

1 & 487120  & 99.4 & 41264  & 94.4 & 91.5 & -5.0\\

2 & 486853  & 99.6 & 42833  & 94.4 & 91.2 & -5.2\\

3 & 486931  & 96.1 & 41320  & 94.4 & 91.5 & -1.7\\

4 & 486901  & 95.2 & 42055  & 95.3 & 91.4 & 0.1 \\

5 & 486753  & 95.6 & 42466  & 95.5 & 91.3 & -0.1\\

6 & 486914  & 99.8 & 41165  & 95.3 & 91.5 & -4.5\\

7 & 486992  & 97.9 & 40674  & 95.3  & 91.6 & -2.6\\

8 & 486789  & 95.3 & 40523  & 95.3 & 91.7 & 0.0\\

9 & 486782  & 94.4 & 41471  & 94.4 & 91.5 & 0.0\\

10 & 486737  & 95.3 & 41642  & 95.3 & 91.4 & 0.0\\

11 & 487124  & 97.6 & 41741  & 95.3 & 91.4 & -2.3\\

12 & 486838  & 96.4 & 42006  & 95.3  & 91.4 & -1.1\\

13 & 486819  & 97.1 & 40762  & 95.3 & 91.6 & -1.8\\

14 & 486844  & 96.8 & 42055  & 95.3 & 91.4 & -1.5\\

15 & 486899  & 96.4 & 40559  & 94.4  & 91.7 & -2.0\\

16 & 486816  & 95.3 & 40741  & 94.4 & 91.6 & -0.9\\

17 & 486844  & 97.6 & 40543  & 94.4 & 91.7 & -3.2\\
\bottomrule   
Average &  &  &  &  & 91.5 & -1.8\\
\end{tabular}  
}
\caption{Compression Statistics with Pruning, Clustering, and Quantization}  
\label{Table VIII}
\end{table}

\section{Conclusion}
\label{sec: conclusion}
This paper studies deep compression techniques for fault diagnosis on the Tennessee Eastman process. In respond to the demand of fast online detection for industrial processes, three compression techniques (pruning, clustering, and quantization) are applied to reduce model size and computational complexity. We have examined a comprehensive list of 7 different combinations of compression techniques. All methods achieve high model compression rates over 64\% while maintaining high fault detection accuracy. The best candidate for fault detection on the Tennessee Eastman chemical process is applying all three techniques, which reduces model sizes by 91.5\% and remains a high accuracy over 94\%. This result leads to smaller storage requirement in production environment, and makes the deployment smoother in real world.

\bibliographystyle{unsrt}

\begin{thebibliography}{1}

\bibitem{Ciresan:2012}
Dan Ciresan, Ueli Meier, and Jürgen Schmidhuber.
\newblock Multi-column deep neural networks for image classification.
\newblock In {\em IN PROCEEDINGS OF THE 25TH IEEE CONFERENCE ON COMPUTER VISION
  AND PATTERN RECOGNITION (CVPR 2012}, pages 3642--3649, 2012.

\bibitem{Socher:2013}
Richard Socher, Alex Perelygin, Jean Wu, Jason Chuang, Christopher~D. Manning,
  Andrew Ng, and Christopher Potts.
\newblock Recursive deep models for semantic compositionality over a sentiment
  treebank.
\newblock In {\em Proceedings of the 2013 Conference on Empirical Methods in
  Natural Language Processing}, pages 1631--1642, Seattle, Washington, USA,
  October 2013. Association for Computational Linguistics.

\bibitem{Chen:2015}
C.~{Chen}, A.~{Seff}, A.~{Kornhauser}, and J.~{Xiao}.
\newblock Deepdriving: Learning affordance for direct perception in autonomous
  driving.
\newblock In {\em 2015 IEEE International Conference on Computer Vision
  (ICCV)}, pages 2722--2730, 2015.

\bibitem{Downs:1993}
J.J. Downs and E.F. Vogel.
\newblock A plant-wide industrial process control problem.
\newblock {\em Computers \& Chemical Engineering}, 17(3):245 -- 255, 1993.
\newblock Industrial challenge problems in process control.

\bibitem{Heo:2018}
Seongmin Heo and Jay~H. Lee.
\newblock Fault detection and classification using artificial neural networks.
\newblock {\em IFAC-PapersOnLine}, 51(18):470 -- 475, 2018.
\newblock 10th IFAC Symposium on Advanced Control of Chemical Processes ADCHEM
  2018.

\bibitem{Chiang:2001}
Leo Chiang, E.~Russell, and Richard Braatz.
\newblock Fault detection and diagnosis in industrial systems.
\newblock {\em Measurement Science and Technology - MEAS SCI TECHNOL}, 12, 10
  2001.

\bibitem{Zhang:2009}
Ying wei Zhang.
\newblock Enhanced statistical analysis of nonlinear processes using kpca, kica
  and svm.
\newblock {\em Chemical Engineering Science}, 64:801--811, 2009.

\end{thebibliography}

\end{document}